\documentclass[sigconf]{acmart}

\AtBeginDocument{%
  \providecommand\BibTeX{{%
    \normalfont B\kern-0.5em{\scshape i\kern-0.25em b}\kern-0.8em\TeX}}}

\setcopyright{none} 
\begin{document}

\settopmatter{printfolios=true}
\title{Improving Human Motion Prediction Through Continual Learning}

\author{Mohammad Samin Yasar}
\affiliation{%
  \institution{University of Virginia}
  \city{Charlottesville, VA 22904}
  \country{USA}
}
\email{msy9an@virginia.edu}

\author{Tariq Iqbal}
\affiliation{%
  \institution{University of Virginia}
  \city{Charlottesville, VA 22904}
  \country{USA}
  }
\email{tiqbal@virginia.edu}

\renewcommand{\shortauthors}{Yasar et al.}

%%
%% The abstract is a short summary of the work to be presented in the article.
\begin{abstract}
Human motion prediction is an essential component for enabling closer human-robot collaboration. The task of accurately predicting human motion is non-trivial. It is compounded by the variability of human motion, both at a skeletal level due to the varying size of humans and at a motion level due to individual movement's idiosyncrasies. These variables make it challenging for learning algorithms to obtain a general representation that is robust to the diverse spatio-temporal patterns of human motion.
In this work, we propose a modular sequence learning approach that allows end-to-end training while also having the flexibility of being fine-tuned. Our approach relies on the diversity of training samples to first learn a robust representation, which can then be fine-tuned in a continual learning setup to predict the motion of new subjects. 
We evaluated the proposed approach by comparing its performance against state-of-the-art baselines. The results suggest that our approach outperforms other methods over all the evaluated temporal horizons, using a small amount of data for fine-tuning. 
The improved performance of our approach opens up the possibility of using continual learning for personalized and reliable motion prediction.
\end{abstract}
\maketitle
\begin{figure}
    \centering
    \includegraphics[width=1\linewidth]{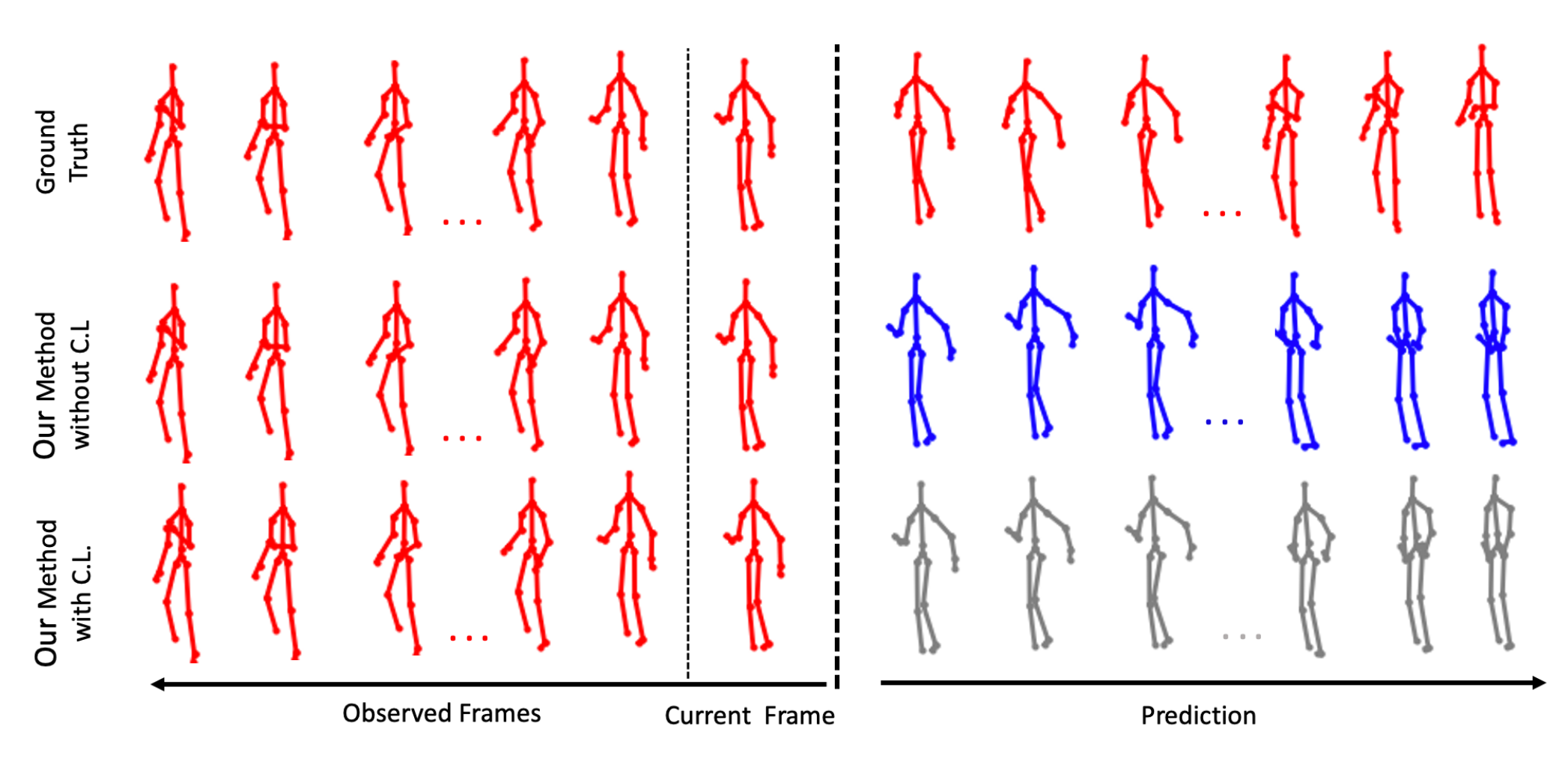}
    \caption{Qualitative performance of different motion prediction methods for walking on UTD-MHAD. The framework trained with Curriculum Learning (C.L.) have predictions that are closer to the ground-truth poses.}

    \label{fig:qualitative comparison}
\end{figure}
\section{Introduction}
%Paragraph 1: What is the problem and why this is a problem
Human motion prediction involves forecasting future human poses given past motion. For enabling efficient Human-Robot Collaboration, a crucial aspect of robot perception is real-time anticipatory modeling of human motion \cite{rudenko2020human,hoffman2019evaluating,Iqbal2017HRI-Bookchapter, Iqbal2016T-RO}. Fluid tasks such as collaborative assembly, handovers, and navigating through moving crowds require combining aspects of perception, representation, and motion analysis to accurately and timely predict probable human motion \cite{biswas2019fast, fosapt, mit_ucsd, islam2020hamlet, islam2020RAL, Iqbal2017RA-L, Iqbal2015TAC}. This would enable the robot to anticipate the human pose and intent and plan accordingly around the human partner without disturbing the natural flow of the human’s motion. However, accurate and timely prediction of human motion remains a non-trivial problem due to the complex and interpersonal nature of human behavior \cite{yasar2021RAL, Iqbal2014Roman}.

%Paragraph 2: Challenges
To address the aperiodic and stochastic nature of human motion, prior work has framed the problem of predicting future poses like that of sequence learning, modeling the spatio-temporal aspect of human motion using Recurrent Neural Networks  \cite{fragkiadaki2015recurrent,martinez2017human, aksan2019structured,yasar2021RAL}. These approaches aim to learn a unified representation from training samples that are expected to generalize for test data. However, generalization comes at the cost of learning individual subtleties of motion, which is crucial for human-robot collaboration. When training these networks, the core assumption is that the given data points are realizations of independent and identically distributed (i.i.d) random variables. However, this assumption is often violated, e.g., when training and test data come from different distributions (dataset bias or domain shift) or the data points are highly interdependent (e.g., when the data exhibits temporal or spatial correlations) \cite{darrell2015machine}. Both these cases are observed in human motion prediction, making it challenging to deploy models trained on benchmark models to the real world. 

%Paragraph 3: How existing approaches try to address this
While generalization at the cost of learning individual preferences is sub-optimal, there is also a need to learn a robust representation over a diverse range of training samples. As such, training and generalizing over a benchmark dataset cannot be discarded and is, in fact, necessary as the first step to accurate motion prediction. Prior work on language modeling has demonstrated the benefit of learning a rich representation on a large training data followed by fine-tuning on a target task \cite{devlin2018bert,yang2019xlnet,brown2020language}. For human motion prediction, this can be posed as a continual learning problem whereby a motion prediction model acquires prior knowledge by observing a large range of human activities. This is followed by fine-tuning its parameters to accurately capture the subtleties of motion prediction for a particular individual. Such a learning setup, however, brings additional challenges to an already non-trivial problem, with prior work on continual learning demonstrating the risk of \emph{catastrophic forgetting} \cite{lesort2020continual, parisi2019continual}.

%Paragraph 4: Your solution overview
To address the challenges mentioned above, we propose a continual learning scheme that can improve human motion prediction accuracy while reducing the risk of catastrophic forgetting. Our framework is modular and is developed to acquire new knowledge and refine existing knowledge based on the new input. In line with prior work on computational neuroscience, which states that the brain must carry out two complementary tasks: generalize across experiences, and retain specific episodic-like events \cite{kumaran2016learning,parisi2019continual}; we utilize a two-phase learning scheme. Our framework aims to learn a robust representation of past observations by training on a benchmark dataset in the first phase. This is achieved by using a modular encoder-decoder architecture with adversarial regularization \cite{yasar2021RAL}, that has state-of-the-art performance on benchmark datasets. In the second phase, we use the representation learning aspect of the framework to condition future poses and fine-tune only the decoder module on new samples in a curriculum learning setup \cite{bengio2009curriculum}. This mitigates the problem of training from scratch while also providing performance gains, both quantitatively over short, mid, and long-term horizons and qualitatively in terms of generating motion that is perceptibly similar to the ground-truth.

% Models trained in isolation on benchmark datasets. Large datasets enable generalization across contexts. Training data might be very different from application scenarios Generalisation comes at the cost of learning individual differences Cumbersome to retrain and update models. 

%Continual Learning Agents acquire and integrate knowledge incrementally about changing environments Data only made available sequentially. Highly sensitive towards changing data conditions Adaptations in learning to avoid forgetting.

% Traditional ML: Robot perception is pre-trained and does not adapt. Most approaches are insensitive to contextual and user-specific attributions. Individual differences in expression are ignored. Continual Learning: Robot uses pre-trained perception and adapts it with each interaction. Interaction help to gather more data.

\section{Problem Formulation}
Formally defined, human motion prediction is the problem of predicting the future human pose over a horizon, given their past pose and any additional contextual information. In this paper, we assume that there is only one agent in the scene. For any particular scenario, the input to our model is the past or \emph{observed} trajectory frames, spanning time $t=1$ to $\tau$,: $\textbf{X} = \{x_1,\ldots,x_\tau \}$. Each frame $x_t \in \mathbb{R}^N$ denotes the $N$-dimensional body pose. $N$ depends on the number of joints in the skeleton, $J$ and the dimension of the joints $D$, where \textcolor{black}{$N =J \times D$}. The expected output of the model is the future trajectory frames over horizon $H$, i.e. the ground truth pose over the horizon $t=\tau+1$ to $\tau + H$:  $\textbf{Y} = \{ y_{\tau+1},\ldots,y_{\tau+H} \}$. 

Our first objective is to learn the underlying representation which would allow the model to generate feasible and accurate human poses $\textbf{\^{Y}} = \{ \hat{y}_{\tau+1},\ldots,\hat{y}_{\tau+H} \}$. 
We assume that future human pose is conditioned on the past observed or generated poses and predict each frame in an auto-regressive manner as formulated below:
\vspace{-0.5em}
\begin{equation}
\label{equation: motion prediction}
p_{\theta} (\textbf{\^{Y}}) = \prod^{\tau+H}_{\delta=\tau+1}  p_{\theta}(\hat{y}_\delta | \hat{y}_{\tau:\delta-1}, x_{1:\tau})
\end{equation}
where the joint distribution is parameterized by $\theta$. 

Next, we use these learned parameters to fine-tune for a specific agent who was not observed during the training phase, using a continual learning setup. Instead of updating all the model parameters, we update a specific module, say the decoder module, with corresponding parameters $\theta^*$. We formulate this as follows, similar to prior work in continual learning \cite{kirkpatrick2017overcoming}:
\begin{equation}
 \log p\;(\theta^*|D ) = \log p\;(D_B|\theta) + \log p\;(\theta|D_A) - \log p\;(D_B)
\end{equation}
where $D_A$ represents the first phase's training data, which involves learning a representation from the large data distribution. $D_B$ represents the second phase's training data, whereby we aim to learn the parameters for a specific human. $logp(\theta|D_A)$ embeds all the prior information learned during the training phase.
%A^{CL}_i : <h_{i-1}, T_{r_i}, M_{i-1}, t_i> -> h_i, M_i

\section{Continual Learning for Human Motion Prediction}
The collective goal of our approach is to accurately predict human motion while being flexible to parameter or architectural updates, given new data. Our overall framework is comprised of an encoder and decoder, trained end-to-end with adversarial regularization on the latent variables, building on top of our prior work \cite{yasar2021RAL}. The encoder aims to learn a rich representation over past trajectories, which the decoder can use to condition its prediction. To improve model stability and robustness of the latent space, we use adversarial regularization through discriminators. This acts as a regularizer during training and can improve the network's stability during parameter updates over new data. We will first describe the overall model for motion prediction and then discuss its flexibility for fine-tuning on a particular agent.

\begin{figure}
    \centering
    \includegraphics[width=\linewidth]{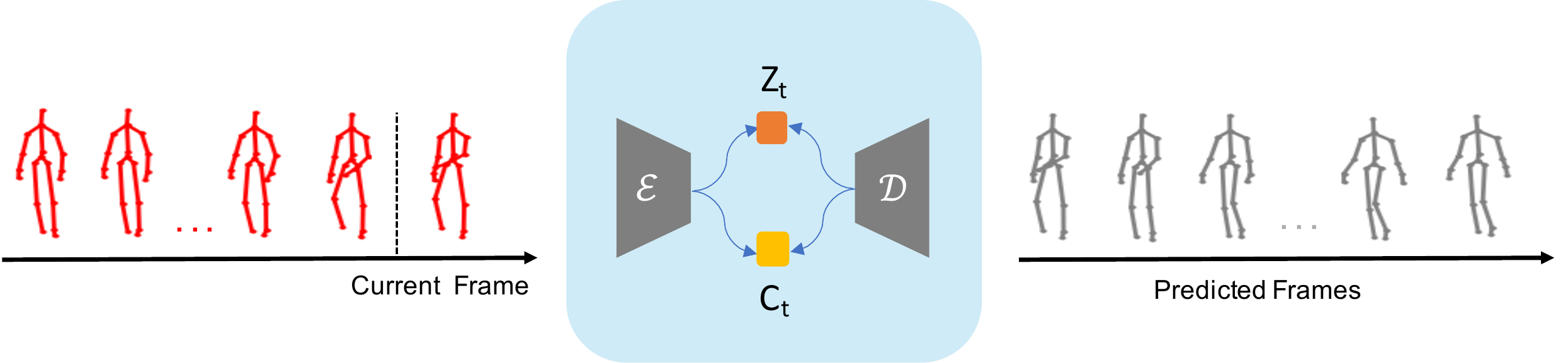}
    \caption{Motion prediction architecture}
    \label{fig:motion prediction}
\end{figure}

\subsection{Overall model for motion prediction}
\textbf{Motion Encoder:}
The encoder learns a representation over the high-dimensional observed trajectory, projecting the input to a low-dimensional latent space. To obtain a rich and more robust representation over the past trajectories, we extract the past velocity and acceleration features along with the provided positional values, in line with prior work on motion prediction \cite{yasar2021RAL}. The velocity and acceleration features are first and second-order derivatives of the position values for each skeleton joint.

For encoding spatio-temporal representation from the position, velocity, and acceleration data, we employ Recurrent Neural Networks, in particular Gated Recurrent Units (GRU). We use unidirectional GRUs, as we wish to predict human motion in real-time. For each stream, the stream-specific GRU aims to extract the spatio-temporal representation that summarizes the input sequence, with the operation formulated as follows:
\begin{equation}
    h_{s,t} = GRU(h_{s,t-1}, x_{s,t}, \phi_s)
\end{equation}
where $s$ represents the specific stream: position, velocity or acceleration, $x_{s,t}$ denotes the input to the GRU at $t$, $h_{s,t-1}$ corresponds to the past hidden state and $\phi_s$ represent the parameters of the GRU. The output from each GRU represents disparate information corresponding to the past trajectory and needs to be fused adaptively. As such, we use a multi-head self-attention mechanism \cite{vaswani2017attention} which is tasked to disentangle and extract relevant stream-specific representation.
\begin{equation}
    \label{equation: attention mechanism}
    \begin{aligned}
        h_t &= Concat(h_{pos,t}; h_{vel,t}; h_{acc,t}); \\
        h_{att,t} &= Attention (h_t; \phi_{att})
    \end{aligned}
\end{equation}
where $h_{att,t}$ is the output of the attention mechanism, and $\phi_{att}$ represents the parameters. The output $h_{att,t}$ is used to obtain the latent representation, which is tasked to characterize the observed trajectory.

\noindent \textbf{Latent Representation:}
The latent representation aims to capture relevant spatial and temporal semantics from the observed data, which can then be used to condition motion prediction. The latent representation is comprised of a continuous random variable and a categorical random variable. The motivation behind using both continuous and categorical variables is to jointly model the continuous aspect of human motion, such as the spatial semantics of a particular activity and the discrete characteristics of human motion such as the class activity or segment.  

To obtain the continuous latent variable $z_t$, the output from the self-attention module is passed through a linear layer. In the case of the categorical latent variable $c_t$, the output from the self-attention module is passed through a linear layer followed by a softmax layer. 
\begin{equation}
    \label{equation: latent variables}
    \begin{split}
     z_t &= Linear(h_{att,t}) \\
     h_{c,t} &= Linear(h_{att,t}) \\
     c_t &= softmax(h_{c,t}) \\
\end{split}
\end{equation}

\noindent \textbf{Adversarial Regularization:}
To enforce a prior on the latent space, we use adversarial learning, similar to the Adversarial Autoencoders (AAE) \cite{makhzani2015adversarial} framework. This serves the purpose of a regularizer as there is a modification to the overall objective function: the objective function now consists of a reconstruction loss and an adversarial loss.  We reason that this helps improve the stability of the overall framework for continual learning, as the parameters are updated based on two competing objectives: the reconstruction loss and the discriminator loss. 

Similar to the GAN \cite{goodfellow2014generative} and AAE \cite{makhzani2015adversarial} setups, the encoder aims to confuse the discriminators by trying to ensure that its output is similar to the aggregated prior. The discriminators are trained to distinguish the true samples generated using a given prior, from the latent space output of the encoder, thus establishing a min-max adversarial game between the networks \cite{goodfellow2014generative, makhzani2015adversarial}.

We use two discriminators, one for the continuous latent variable and the other for the categorical latent variable. The discriminators compute the probability that a point $z_t$ or $c_t$ is a sample from the prior distribution that we are trying to model (positive samples), or from the latent space (negative sample). 
%The discriminator loss, which is high if the generated sample from the encoder is coming from a different distribution compared to the prior, is used to update the parameters of the encoder, thus enforcing it to produce samples similar to the prior, while also providing a robust representation for generating future poses to the decoder. 
We use a Gaussian prior for continuous latent variables and a uniform distribution prior for categorical latent variables. 

\noindent \textbf{Decoder:}
The decoder uses the latent representation and the past generated pose to predict the future pose for each time step. It is auto-regressive, i.e., it uses the output of the previous timestep to predict the current pose and has only one stream: position as the expected output is future joint positions of the human. 

The input to the decoder is the latent representation: $z_t$ and $c_t$ and the past generated pose, or the seed pose at time $t$ if it is predicting the first time-step, $t+1$. This is then passed to an attention mechanism that allows the decoder to adaptively condition its output on the latent variables that provide long-term information over the observed frames and the immediate generated frame. The output from this attention mechanism is next passed to a GRU cell, similar to the one at the encoder. This is followed by a Structured Prediction Layer (SPL) \cite{aksan2019structured}, which predicts each joint hierarchically following a skeleton tree, thus allowing the decoder to enforce structural prior on its final output. The operations at the decoder are formulated as follows: 
\begin{equation}
    \begin{split}
    p_t &= Concat(z_t, c_t, h_{dec, t-1}) \\
    p_{att,t} &= Attention(p_{t}, \phi_{att}) \\
    h_{dec,t} &= GRU(S_{t-1}, p_{att,t}, \phi_{pos})    \\
    S_{t+1} &= \gamma(h_{dec,t}) 
    \end{split}
    \label{equation: decoder}
\end{equation}
\subsection{Curriculum learning for the decoder:} 
The encoder-decoder architecture with adversarial regularization is trained to convergence on the training set. This training is followed by providing the overall architecture with unseen but small samples of motion data. This aims to relax the i.i.d assumption of the training procedure as our framework now has access to limited motion samples of the agent that it is trying to model. 

Our choice of continual learning scheme is the curriculum learning setup \cite{bengio2009curriculum}, whereby we first train the network on a comparatively simpler task of representation learning, followed by a relatively difficult task of fine-tuning its parameters for a specific human subject. Our implementation is based on findings in connectionist models \cite{richardson2008critical, thomas2006computational}, in particular self-organizing maps which reduce the levels of functional plasticity (i.e., ability to acquire knowledge in neural networks) through a two-phase training of the topographic neural map \cite{kohonen1997exploration, kohonen1982self}. The first phase is the organization phase, where the neural network is trained with a high learning rate and large spatial neighborhood size, allowing the network to reach an initial rough topological organization. The second phase is referred to as the tuning phase, where the learning rate and the neighborhood size are iteratively reduced for fine-tuning \cite{parisi2019continual}. We aim to adopt these findings to a sequence learning framework.   

Following prior work on developmental and curriculum learning \cite{bengio2009curriculum, graves2016hybrid}, we fine-tuned the architecture on the new data. We adopt techniques that will allow us to retain previous knowledge and avoid catastrophic forgetting during fine-tuning. In particular, we rely on \emph{discriminative fine-tuning} \cite{howard2018universal}, whereby we fine-tune only the decoder network at a different learning rate while freezing the encoder and the discriminator networks. 

\begin{table*}[t]
    \centering
    \caption{MSE (in cm\textsuperscript{2}) comparison of fune-tuning vs no fine-tuning on UTD-MHAD for different test subjects) (Lower is better)}
    \resizebox{\textwidth}{!}{
    \begin{tabular}{c| c c c c c c| c c c c c c| c c c c c c}
        \hline
         & \multicolumn{6}{c|}{Subject 2} & \multicolumn{6}{c|}{Subject 4} & \multicolumn{6}{c}{Subject 6} \\
        \hline
         Frames  & 2 & 4 & 8 & 10 & 13 & 15 & 2 & 4 & 8 & 10 & 13 & 15 & 2 & 4 & 8 & 10 & 13 & 15 \\
        \hline
        Zero-Velocity & 11.20 & 27.37 & 66.85 & 86.23 & 112.39 & 127.64 
        & 13.11 & 32.33 & 77.22 & 97.80 & 123.99 & 138.53 
        & 10.37 & 25.56 & 64.62 & 85.29 & 115.33 & 135.17
        \\
         Our method without Curriculum-Learning & \textbf{5.41} & \textbf{14.75} & 33.87 & 41.89 & 51.33 & 56.12 
        & 7.78 & 18.26 & 41.66 & 51.46 & 62.67 & 68.1
        & 6.99 & 16.18 & 38.51 & 49.40 & 63.7 & 71.68
         \\ 
         \textbf{Our method with Curriculum-Learning} & 7.62 & 16.17 & \textbf{32.35} & \textbf{38.63} & \textbf{45.72} & \textbf{49.13}  &
         \textbf{6.98} & \textbf{14.79} & \textbf{30.29} & \textbf{36.51} & \textbf{43.81} & \textbf{47.61} &
         \textbf{6.51} & \textbf{13.44} & \textbf{27.49} & \textbf{33.66} & \textbf{41.74} & \textbf{46.41}\\ 
         \hline
    \end{tabular}
    }
    \label{tab:Finetuning all methods}
\end{table*}

\noindent \textbf{Fine-tuning the decoder:}
In line with equation~\ref{equation: latent variables}, the input to the model is the sequence of observed poses: $\textbf{X} = \{x_1,\ldots,x_\tau \}$, with the output being of the encoder being $z_\tau$ and $c_\tau$. However, instead of imposing a prior on the latent space and training the encoder-decoder end-to-end, we only update the decoder's parameters. We also use a lower learning rate and rely on a small number of training samples to improve model stability and reduce the likelihood of catastrophic forgetting. 

We leverage the representation learning capability of the framework that it attained when training on a large and diverse dataset. The encoder network is tasked to provide a representation summarizing the past observation that is used by the decoder to condition its prediction, similar to equation \ref{equation: decoder}. The pre-trained weights from the training set are used to initialize the overall architecture and act as prior knowledge. The decoder weights are updated based on the reconstruction loss on the new data.

\section{Experimental Setup}
\subsection{Dataset}
We evaluated the performance of our approach on the widely used human-activity dataset: UTD-MHAD \cite{chen2015utd}. The dataset comprises 27 action classes covering activities from hand gestures to training exercises and daily activities, thus providing relevant activities for human-robot collaboration. Each activity was performed by 8 different subjects, with each subject repeating the activity 4 times. In our experiments, we use only Skeleton data for predicting human motion, following previous work in this domain \cite{jain2016structural,fragkiadaki2015recurrent, martinez2017human,butepage2018anticipating,aksan2019structured}, and considered each of the 20 provided joints. For all experiments, the model predicted output for the next 15 frames, using observation over the past 15 frames.

\subsection{Generalized Representation Learning}
We used the cross-subject evaluation scheme, training and validating on odd-numbered subjects for the first phase, thus providing the framework with a large training sample and maximizing the likelihood of encountering diverse demonstrations. To evaluate the performance, we hold out a section of the data for the validation set and early stopping. This reduces the likelihood of overfitting on the training data while also provide the mechanism for stopping the training procedure. 

\subsection{Curriculum Learning for a specific subject}
Having learned a generalized representation, the second phase involved training the framework in a curriculum learning setup. Here, the experiments are conducted on a particular held-out even-numbered subject. We fine-tuned only the decoder using a reduced learning rate, with the encoder weights initialized from the first phase. As each subject has 4 trials, we trained on one trial and tested on the other 3 trials.

\subsection{State-of-the-art method and baseline}
For evaluating the efficacy of our curriculum learning setup, we compared against a non-curriculum learning framework, and the zero-velocity baseline \cite{martinez2017human}. The first benchmark \cite{yasar2021RAL} is comprised of an encoder-decoder framework, with adversarial regularization, but with no provision for curriculum learning. The zero-velocity baseline assumes that all the future predictions are identical to the last observed pose and is challenging to outperform for short-term prediction \cite{martinez2017human, aksan2019structured}. It also allows us to gauge the movement dynamics, with a lower MSE for zero-velocity suggesting less movement and vice-versa for higher MSE.

\subsection{Evaluation Metric}
We evaluated the performance of all models using the Mean Squared Error (MSE), which is the $l_2$ distance between the ground-truth and the predicted poses at each timestep, averaged over the number of joints and sequence length, in line with prior work \cite{butepage2017deep, adeli2020socially, butepage2019imitating, yasar2021RAL}. The MSE is calculated as:

\begin{equation}
    \mathcal{L (X, \hat{X})} =  \frac{1}{T.K}\sum^T_{t=1} \sum^K_{i=1} (x_{t}^{i} - \hat{x}_{t}^{i})^2 \\
\end{equation}

where, T and K are the total number of frame and joints respectively.

\section{Results and Discussion}
\textbf{Results:} We present the results of all approaches on the UTD-MHAD on table \ref{tab:Finetuning all methods}. 
We report the performance of all approaches at distinct frame intervals to circumvent the problem of frame drops during data collection and subsequent evaluation \cite{yasar2021RAL}. Our frame intervals aim to evaluate all models on short (2 \& 4), mid (8 \& 10), and long-term motion prediction (13 \& 15). Table I depicts the performance of all approaches with respect to the test subjects. The results in Table I suggest that fine-turning the framework allows it to outperforms all other methods and the zero-velocity baseline for short, mid, and long-term prediction.  

\noindent \textbf{Discussion:} Our proposed approach outperformed the prior state-of-the-art approach and baseline both quantitatively by having lower MSE and qualitatively in terms of generating motion closer to the ground-truth pose (see Fig. \ref{fig:qualitative comparison}). This shows the benefit of the curriculum learning approach while also suggesting that our overall framework is robust to catastrophic forgetting. The performance gain is especially significant over the mid and long-term, as the decoder is trained to learn the spatial-temporal movement pattern of a specific subject and can generate the future pose with higher accuracy. 

Using a curriculum-learning setup, albeit on a small training sample, allows the framework to capture individual human motion subtleties, as seen by the lower MSE, particularly over the mid and long-term horizons. The approach is particularly useful when there is significant movement over the given horizon, as seen for Subjects 4 and 6 (table \ref{tab:Finetuning all methods}), who have higher MSE loss on the zero-velocity baseline. For Subject 2, there is overall less movement as seen by the zero-velocity MSE loss, and hence the performance gain is not significant over the mid and long-term and even worse over the short-term. Overall, the results are particularly promising as we did not fine-tune the encoder, instead only focusing on the decoder. Further improvement can be attained by fine-tuning the encoder.

\section{Conclusion}
In this work, we present a curriculum learning approach that opens the possibility of continual learning for human motion prediction, especially if the model is especially \emph{deployed in the wild}. Our framework first learns a general representation over diverse training samples before fine-tuning on a target human subject. Our experiments suggest the feasibility of curriculum learning with performance gains over non-curriculum learning approaches. Future work will focus on fine-tuning to activities and domains that were not observed in training in a zero-shot learning setup.

\section{Acknowledgement}
This work is supported by the CCAM Innovation Award. The authors thank Tim Bakker, Tomonari Furukawa, and Qing Chang for their support.

\bibliographystyle{ACM-Reference-Format}
\bibliography{general.bib, architectures.bib, forecasting.bib, control.bib, workflowanalysis.bib}

\appendix

\end{document}